\pdfoutput=1
\documentclass[10pt, a4paper]{article}

\usepackage{lrec}
\usepackage{times}
\usepackage{url}
\usepackage{latexsym}
\usepackage{makecell}
\usepackage{booktabs}
\usepackage{ltxtable} 
\usepackage{tabularx}
\usepackage{amsmath}
\usepackage{soul}
\usepackage{hyperref}
\usepackage[usenames, dvipsnames]{xcolor}

\DeclareRobustCommand{\hlcyan}[1]{{\sethlcolor{cyan}\hl{#1}}}

\DeclareRobustCommand{\hlgreen}[1]{{\sethlcolor{ForestGreen}\hl{#1}}}
\DeclareRobustCommand{\hlgray}[1]{{\sethlcolor{gray}\hl{#1}}}

\DeclareMathOperator*{\argmax}{arg\,max}
\usepackage{longtable}

\title{\textbf{Data-driven Summarization of Scientific Articles}}

\name{Nikola I. Nikolov, Michael Pfeiffer, Richard H.R. Hahnloser}

\address{Institute of Neuroinformatics, University of Z{\"u}rich and ETH Z{\"u}rich, Switzerland \\
   {\tt \{niniko, pfeiffer, rich\}@ini.ethz.ch}}

\usepackage[acronym]{glossaries}
\usepackage{graphicx}
\makenoidxglossaries
\newacronym{nlp}{NLP}{Natural Language Processing}
\newacronym{lstm}{LSTM}{Long Short-Term Memory}
\newacronym{gru}{GRU}{Gated Recurrent Unit}
\newacronym{birnn}{BiRNN}{Bidirectional Recurrent Neural Network}
\newacronym{mlp}{MLP}{Multi-Layered Perceptron}
\newacronym{mt}{MT}{Machine Translation}
\newacronym{nmt}{NMT}{Neural Machine Translation}
\newacronym{rnn}{RNN}{Recurrent Neural Network}
\newacronym{cnn}{CNN}{Convolutional Neural Network}
\newacronym{eos}{EOS}{End-of-sequence}
\newacronym{rnnlm}{RNNLM}{Recurrent Neural Network Language Model}
\newacronym{nnlm}{NNLM}{Neural Network Language Model}
\newacronym{abs}{ABS}{Attention-based Summarization}
\newacronym{rst}{RST}{Rhetorical structure theory}
\newacronym{bptt}{BPTT}{Back-propagation through time}
\newacronym{hrnns}{HieRNNsearch}{Hierarchical RNNsearch}
\newacronym{rouge}{ROUGE}{Recall-Oriented Understudy for Gisting Evaluation}
\newacronym{duc}{DUC}{Document Understanding Conference}
\newacronym{bleu}{BLEU}{Bilingual Evaluation Understudy}

\usepackage{booktabs}
\usepackage{ltxtable} 

\usepackage{subcaption}
\usepackage[export]{adjustbox}
\usepackage{placeins}
\usepackage{float}

\abstract{
Data-driven approaches to sequence-to-sequence modelling have been successfully applied to short text summarization of news articles. Such models are typically trained on input-summary pairs consisting of only a single or a few sentences, partially due to limited availability of multi-sentence training data. Here, we propose to use scientific articles as a new milestone for text summarization: large-scale training data come almost for free with two types of high-quality summaries at different levels - the title and the abstract. We generate two novel multi-sentence summarization datasets from scientific articles and test the suitability of a wide range of existing extractive and abstractive neural network-based summarization approaches. Our analysis demonstrates that scientific papers are suitable for data-driven text summarization. Our results could serve as valuable benchmarks for scaling sequence-to-sequence models to very long sequences.}

\begin{document}

\maketitleabstract

\captionsetup[table]{skip=0pt}

\section{Introduction}

The goal of automatic text summarization is to produce a shorter, informative version of an input text. While extractive summarization only consists of selecting important sentences from the input, abstractive summarization \textit{generates} content without explicitly re-using whole sentences \cite{nenkova2011automatic}. Text summarization is an area with much promise in today's age of information overflow. In the domain of scientific literature, the rate of publications grows exponentially \cite{hunter2006biomedical}, which calls for efficient automatic summarization tools. 

Recent state-of-the-art summarization methods learn to summarize in a data-driven way, relying on large collections of input-summary training examples. The majority of previous work focused on short summarization of news articles, such as  to generate a title \cite{rushneural,nallapati2016sequence}. One major challenge is to scale these methods to process long input/output sequence pairs. Currently, availability of large-scale high-quality training data is scarce. 

In this paper, we explore the suitability of \textbf{scientific journal articles} as a new benchmark for data-driven text summarization. The typical well-structured format of scientific papers makes them an interesting challenge, and provides plenty of freely available training data, because every article comes with a summary in the form of its abstract, and, in even more compressed form, its title. We make a fist step towards summarization of whole scientific articles, by composing two novel large datasets for \textbf{scientific summarization}: title-abstract pairs (\textit{title-gen}), composed of 5 million papers in the biomedical domain, and abstract-body pairs (\textit{abstract-gen}) composed of 900k papers\footnote{Both datasets are available at \url{https://github.com/ninikolov/data-driven-summarization}, including versions with and without preprocessing.}. The second dataset is particularly challenging, because it is intended for summarizing the full body of the paper in terms of the abstract (the lengths of input/output sequences are substantially longer than what has been considered so far in previous research, see Table \ref{table:datasets}). 

We evaluate a range of existing state-of-the-art approaches on these datasets: extractive approaches based on word embeddings, as well as word, subword, and character-level encoder-decoder models that use recurrent as well as convolutional modules. We perform a quantitative and qualitative analysis of the models' outputs. 

\section{Background}

\subsection{Extractive Summarization}\label{extractive}

Given an input document consisting of $T_s$ sentences $\pmb{s} = \{s_1,...,s_{T_s}\}$, the goal of \textbf{extractive summarization} is to select the $K$ most salient sentences as the output summary. Extractive summarization typically involves a \textbf{sentence representation module} $e$, that represents each input sentence $s_i$ in a common space as $r_i = e(s_i)$, e.g. as a vector of real numbers; as well as a \textbf{ranking module} $score$, that weights the salience $w_i = score(r_i)$ of each sentence. A typical approach to unsupervised extractive summarization is to implement $w_i$ as the similarity between $r_i$ and a document representation (or a \textbf{document centroid}) $r_d = e(d)$ \cite{radev2004centroid}. Alternatively, one can compute $w_i$ as the \textbf{sentence centrality}, which is an adjacency-based measure of sentence importance \cite{erkan2004lexrank}. 

In this work, we propose two simple unsupervised baselines for extractive summarization, both of which rely on word embeddings \cite{mikolov2013efficient}. The first, \textit{tfidf-emb}, represents each sentence in the input document as the weighted sum of its constituent word embeddings, similar to \cite{rossiello2017centroid}: 

\begin{equation}
    r_i=e(s_i) = \frac{1}{t_i}\sum_{x \in s_i}t(x)\cdot E(x),
\end{equation}

where $E(x)$ is the embedding of word $x$, $t(x)$ is an (optional) weighting function that weighs the importance of a word, and $t_i=\sum_{x\in s_i} t(x)$ is a normalization factor. As a weighing function, we use the term-frequency inverse document frequency (TF-IDF) score, similar to \cite{brokos2016using}. Each sentence embedding $r_i$ can then be ranked by computing its cosine similarity $sim(r_d, r_i)$ to a document centroid $r_d$, computed similarly as $r_i$. The summary consists of the top $K$ sentences with embeddings most similar to the document embedding.

The second baseline, \textit{rwmd-rank}, ranks the salience of a sentence in terms of its similarity to all the other sentences in the document. All similarities are stored in an intra-sentence similarity matrix $W$. We use the Relaxed Word Mover's Distance (RWMD) to compute this matrix \cite{kusner2015word}: 

\begin{equation}
    \begin{split}
        W_{ij} = rwmd(s_i, s_j) = \max (rwmd_p(s_i, s_j), rwmd_p(s_j, s_i)) \\
        rwmd_{p}(s_i, s_j) = \sum_{x\in s_i} \min_{x' \in s_j} dist(E(x), E(x')), 
    \end{split}
\end{equation}

where $s_i$ and $s_j$ are two sentences and $dist(E(x), E(x'))$ is the Euclidean distance between the embeddings of words $x$ and $x'$ in the sentences. To rank the sentences, we apply the graph-based method from the LexRank system \cite{erkan2004lexrank}. LexRank represents the input as a highly connected graph, in which vertices represent sentences, and edges between sentences are assigned weights equal to their similarity from $W$. The centrality of a sentence is then computed using the PageRank algorithm \cite{page1999pagerank}. 

\subsection{Abstractive Summarization}\label{sec:abstractive}

Given an input sequence of $T_x$ words $\pmb{x} = \{x_1,...,x_{T_x}\}$ coming from a fixed-length input vocabulary $V_x$ of size $K_x$, the goal of abstractive summarization is to produce a condensed sequence of $T_y$ summary words $\pmb{y} = \{y_1,...,y_{T_y}\}$ from a summarization vocabulary $V_y$ of size $K_y$, where $T_x \gg T_y$. Abstractive summarization is a structured prediction problem that can be solved by learning a probabilistic mapping $p(\pmb{y}|\pmb{x}, \theta)$ for the summary $\pmb{y}$, given the input sequence $\pmb{x}$ \cite{dietterich2008structured}:

\begin{equation}\label{eq:mapping}
\begin{split}    
  p(\pmb{y}|\pmb{x}, \theta) = \prod^{T_y}_{i} p(y_i|\{y_0,...,y_{i-1}\}, \pmb{x}, \theta).
\end{split}
\end{equation}

The \textbf{encoder-decoder architecture} 
is a recently proposed general framework for structured prediction \cite{cho2015describing}, in which the distribution $\argmax_{\pmb{y}}p(\pmb{y}|\pmb{x}, \theta)$ is learned using two neural networks: an \textbf{encoder} network $e$, which produces intermediate representations of the input, and a \textbf{decoder} language modelling network $d$, which generates the target summary. The decoder is conditioned on a context vector $c$, which is recomputed from the encoded representation at each decoding step. The encoder-decoder was first implemented using Recurrent Neural Networks (RNNs) \cite{sutskever2014sequence,cho2014learning} that process the input sequentially. Recent studies have shown that convolutional neural networks (CNNs) \cite{Lecun1998CNN} can outperform RNNs in sequence transduction tasks \cite{Kalchbrenner2016Bytenet,gehring2017convolutional}. Unlike RNNs, CNNs can be efficiently implemented on parallel GPU hardware. This advantage is particularly important when working with very long input and output sequences, such as whole paragraphs or documents. CNNs create hierarchical representations over the input in which lower layers operate on nearby elements and higher layers implement increasing levels of abstraction. 

In this work, we investigate the performance of three existing systems that operate on different levels of sequence granularity. The first, \textit{lstm}, is a recurrent Long Short Term Memory (LSTM) encoder-decoder model \cite{sutskever2014sequence} with an attention mechanism \cite{bahdanau2014neural} that operates on the word level, processing the input sequentially. The second system, \textit{fconv}, is a convolutional encoder-decoder model from \cite{gehring2017convolutional}. \textit{fconv} works on the subword level and segments words into smaller units using the byte pair encoding scheme. Using subword units improves the generation quality when dealing with rare or unknown words \cite{sennrich2015neural}. The third system, \textit{c2c}, is a character-level encoder-decoder model from \cite{lee2016fully} that models $\pmb{x}$ and $\pmb{y}$ as individual characters, with no explicit segmentation between tokens. \textit{c2c} first builds representations of groups of characters in the input using a series of convolutional layers. It then applies a recurrent encoder-decoder, similar to the \textit{lstm} system. 

\subsection{Scientific Articles}

Previous research on summarization of scientific articles has focused almost exclusively on extractive methods \cite{nenkova2011automatic}. In \cite{lloret2013compendium}, the authors develop an unsupervised system for abstract generation of biomedical papers, that first selects relevant content from the body, following which it performs an abstractive information fusion step. More recently, \cite{kim2016towards} consider the problem of supervised generation of sentence-level summaries for each paragraph of the introduction of a paper. They construct a training dataset of computer science papers from arXiv, selecting the most informative sentence as the summary of each paragraph using the Jaccard similarity. Thus, their target summary is fully contained in the input. In \cite{collins2017supervised}, they develop a supervised extractive summarization framework which they apply to a dataset of $10k$ computer science papers. To the best of our knowledge, our work is the first on abstractive title generation of scientific articles, and is the first to consider supervised generation of the absctract directly from the full body of the paper. The datasets we utilize here are also substantially larger than in previous work on scientific summarization. 

Scientific articles are potentially more challenging to summarize than news articles because of their compact, inexplicit discourse style \cite{biber2010challenging}. While the events described by news headlines frequently recur in related articles, a scientific title focuses on the unique contribution that sets a paper apart from previous research \cite{teufel2002summarizing}. Furthermore, while the first two sentences of a news article are typically sufficiently informative to generate its headline \cite{nallapati2016sequence,teufel2002summarizing}, the first sentences of the abstract or introduction of a paper typically contain background information on the research topic. Constructing a good scientific title thus requires understanding and integrating concepts from multiple sentences of the abstract. 

\section{Datasets}\label{sec:dataset}

\begin{table*}[]
\centering
\caption{
\small{Statistics (mean and standard deviation) of the two scientific summarization datasets: \textit{title-gen} and \textit{abstract-gen}.  Token/sentence counts are computed with NLTK.}}
\label{table:datasets}
\parbox{.45\linewidth}{
\begin{tabular}{|c|cc|}
\hline 
\textit{title-gen}               & \textit{Abstract} & \multicolumn{1}{c|}{\textit{Title}} \\ \hline
\textbf{Token count}  & $245 \pm 54$          & \multicolumn{1}{c|}{$15\pm4$}         \\ \hline
\textbf{Sentence count} & $14\pm4$         & \multicolumn{1}{c|}{1}              \\ \hline
\textbf{Sent. token count} & 26 $\pm$ 14 & - 
\\ \hline
\textbf{Overlap}        & \multicolumn{2}{c|}{$73\% \pm 18\%$}\\ \hline 
\textbf{Repeat} & $44\% \pm 11\%$ & -  \\ \hline 
\textbf{Size (tr/val/test)}        & \multicolumn{2}{c|}{$5'000'000/6844/6935$} \\ \hline 

\end{tabular}
}
\parbox{.45\linewidth}{
\begin{tabular}{|c|cc|}
\hline
\textit{abstract-gen}               & \textit{Body} & \multicolumn{1}{c|}{\textit{Abstract}} \\ \hline
\textbf{Token count}     & $4600 \pm 1987$          & \multicolumn{1}{c|}{$254 \pm 54$}         \\ \hline
\textbf{Sentence count} & $172 \pm 78$         & \multicolumn{1}{c|}{$10 \pm 3$}              
\\ \hline
\textbf{Sent. token count} & 26 $\pm$ 17 & 26 $\pm$ 14
\\ \hline
\textbf{Overlap}        & \multicolumn{2}{c|}{$68\% \pm 10\%$}\\ \hline 
\textbf{Repeat} & $74\% \pm 7\%$ & $44\% \pm 11\%$ \\ \hline 
\textbf{Size (tr/val/test)}        & \multicolumn{2}{c|}{$893'835/10'916/10'812$} \\ \hline 
\end{tabular}
}
\end{table*}

To investigate the performance of encoder-decoder neural networks as generative models of scientific text, we constructed two novel datasets for scientific summarization. For \textit{title-gen} we used MEDLINE\footnote{\url{https://nlm.nih.gov/databases/download/pubmed_medline.html}}, whereas for \textit{abstract-gen} we used the PubMed open access subset\footnote{\url{https://ncbi.nlm.nih.gov/pmc/tools/openftlist}}. MEDLINE contains scientific metadata in XML format of $\sim25$ million papers in the biomedical domain, whereas the PubMed open access subset contains metadata and full text of $\sim1.5$ million papers.

We processed the XML files\footnote{We use \url{https://titipata.github.io/pubmed_parser}.} to pair the abstract of a paper to its title (\textit{title-gen} dataset) or the full body (\textit{abstract-gen}), skipping any figures, tables or section headings in the body. We then apply several preprocessing steps from the MOSES statistical machine translation pipeline\footnote{\url{github.com/moses-smt/mosesdecoder}}, including tokenization and conversion to lowercase. Any URLs were removed, all numbers replaced with \#, and any pairs with abstract lengths not in the range of 150-370 tokens, title lengths not within 6-25 tokens, and body lengths not within 700-10000 tokens were excluded. 

The \textit{Overlap} $o(\mathbf{x}, \mathbf{y})$
$= \frac{|\{\mathbf{y}\} 	\cap \{\mathbf{x}\}|}{|\{\mathbf{y}\}|}$ 
is the fraction of unique output (summary) tokens $\mathbf{y}$ that overlap with an input token $\mathbf{x}$ (excluding punctuation and stop words). As can be seen in Table \ref{table:datasets}, the overlaps are large in our datasets, indicating frequent reuse of words. The \textit{Repeat}  $e(\mathbf{s})=\frac{\sum_i o(\overline{\mathbf s_i}, s_{i})}{|\mathbf{s}|}$  is the average overlap of each sentence $s_{i}$ in a text with the remainder of the text (where $\overline{\mathbf s_i}$ denotes the complement of sentence $s_{i}$). \textit{Repeat} measures the redundancy of content within a text: a high value indicates frequent repetition of content. Whereas in abstracts there are only moderate levels of repetition, in the bodies the repetition rates are much higher, possibly because concepts and ideas are reiterated in multiple sections of the paper. 

\section{Evaluation Set-up}

We evaluated the performance of several state-of-the-art approaches on our scientific summarization datasets. The extractive systems we consider are: \textit{lead}, \textit{lexrank}, \textit{tfidf-emb}, and \textit{rwmd-rank}. The \textit{lead} baseline returns the first sentence of the abstract for \textit{title-gen}, or the first 10 sentences of the body for \textit{abstract-gen}. \textit{lexrank} \cite{erkan2004lexrank} is a graph-based centrality approach frequently used as a baseline in the literature. \textit{emb-tfidf} uses sentence embeddings\footnote{We use the best-performing Word2Vec model from \cite{chiu2016train}, which is trained on PubMed and MEDLINE.} to select the most salient sentences from the input, while \textit{rwmd-rank} uses the Relaxed Word Mover's Distance (as described in Section \ref{extractive}). \textit{oracle} estimates an upper bound for the extractive summarization task by finding the most similar sentence in the input document for each sentence in the original summary. We use the Relaxed Word Mover's Distance to compute the output of the oracle. 

The abstractive systems we consider are: \textit{lstm}, \textit{fconv}, and \textit{c2c}, described in Section \ref{sec:abstractive}. For \textit{lstm}, we set the input/output vocabularies to $80 000$, use two LSTM layers of 1000 hidden units each, and word embedding of dimension 500 (we found no improvement from additionally increasing the size of this model). For \textit{c2c} and \textit{fconv}, we use the default hyper-parameters that come with the public implementations provided by the authors of the systems. The \textit{title-gen} \textit{lstm}, \textit{c2c}, and \textit{fconv} were trained for 11, 8, and 20 epochs, respectively, until convergence. 

We were unable to train \textit{lstm} and \textit{c2c} on \textit{abstract-gen} because of the very high memory and time requirements associated with the recurrent layers in these models. We found \textit{fconv} to be much more efficient to train, and we succeeded in training a default model for 17 epochs. For \textit{title-gen}, we used beam search with beam size $20$, while for \textit{abstract-gen} we found a beam size of $5$ to perform better. 

\subsection{Quantitative Evaluation}

\begin{table*}
\centering
\caption{\small{Metric results for the \textit{title-gen} dataset. R-1, R-2, R-L represent the ROUGE-1/2/L metrics.}}
\label{table:title}
\begin{tabular}{|c|c|c|c|c|c|c|c|}
\hline
\textbf{Model} & \textbf{R-1} & \textbf{R-2} & \textbf{R-L} & \textbf{METEOR} & \textbf{Overlap} & \textbf{Token count} \\ \Xhline{2\arrayrulewidth}
\textit{oracle} & 0.386 & 0.184 & 0.308 & 0.146 & \multicolumn{1}{c|}{-} & 29 $\pm$ 14 \\ \Xhline{2\arrayrulewidth}
\textit{lead-1} & 0.218 & 0.061 & 0.169 & 0.077 & \multicolumn{1}{c|}{-} & 28 $\pm$ 14  \\ \hline 
\textit{lexrank} & 0.26 & 0.089 & 0.201 & 0.089 & \multicolumn{1}{c|}{-} & 32 $\pm$ 14 \\ \hline 
\textit{emb-tfidf} & 0.252 & 0.081 & 0.193 & 0.082 & \multicolumn{1}{c|}{-} & 35 $\pm$ 17 \\ \hline
\textit{rwmd-rank} & 0.311 & 0.13 & 0.245 & 0.116 & \multicolumn{1}{c|}{-} & 28 $\pm$ 13 \\ \Xhline{2\arrayrulewidth}
\textit{lstm} & 0.375  & 0.173 & 0.329 & 0.204 & \multicolumn{1}{c|}{78\% $\pm$ 20\%} & 12 $\pm$ 3 \\ \hline 
\textit{c2c} & \textbf{0.479} & 0.264 & \textbf{0.418} & 0.237 &
\multicolumn{1}{c|}{93\% $\pm$ 10\%} & 14 $\pm$ 4 \\ \hline 
\textit{fconv} & 0.463 & \textbf{0.277} & 0.412 & \textbf{0.27} &
\multicolumn{1}{c|}{95\% $\pm$ 9\%} & 15 $\pm$ 7
\\ \hline 
\end{tabular}
\end{table*}
\begin{table*}
\centering
\caption{\small{Metric results for the \textit{abstract-gen} dataset. R-1, R-2, R-L represent the ROUGE-1/2/L metrics.}}
\label{table:abstract}
\begin{tabular}{|c|c|c|c|c|c|c|c|}
\hline
\textbf{Model} & \textbf{R-1} & \textbf{R-2} & \textbf{R-L} & \textbf{METEOR} & \textbf{Overlap} & \textbf{Repeat} & \textbf{Token count} \\ \Xhline{2\arrayrulewidth} 
\textit{oracle} & 0.558 & 0.266 & 0.316 & 0.214 & - & \multicolumn{1}{c|}{42\% $\pm$ 10\%} & 327 $\pm$ 99 \\ \Xhline{2\arrayrulewidth}
\textit{lead-10} & 0.385 & 0.111 & 0.18 & 0.138 & - & \multicolumn{1}{c|}{20\% $\pm$ 4\%} & 312 $\pm$ 88 \\ \hline 
\textit{lexrank} & 0.45 & \textbf{0.163} & 0.213 & 0.157 & - & \multicolumn{1}{c|}{52\% $\pm$ 10\%} & 404 $\pm$ 131 \\ \hline 
\textit{emb-tfidf} & 0.445 & 0.159 & 0.216 & 0.159 & - & \multicolumn{1}{c|}{52\% $\pm$ 10\%} & 369 $\pm$ 117 \\ \hline 
\textit{rwmd-rank} & \textbf{0.454} & 0.159 & \textbf{0.216} & 0.167 & - & \multicolumn{1}{c|}{50\% $\pm$ 10\%} & 344 $\pm$ 93 \\ \Xhline{2\arrayrulewidth}
\textit{fconv} & 0.354 & 0.131 & 0.209 & \textbf{0.212} & 98\% $\pm$ 2\% & \multicolumn{1}{c|}{52\%  $\pm$ 28\%} & 194 $\pm$ 15 \\ \hline 
\end{tabular}
\end{table*}

In Tables \ref{table:title} and \ref{table:abstract}, we evaluate our approaches using the ROUGE metric \cite{lin2004rouge}, which is a recall-based metric frequently used for summarization, and METEOR \cite{denkowski:lavie:meteor-wmt:2014}, which is a precision-based metric for machine translation. Overlap can be interpreted as the tendency of the model to directly copy input content instead of generating novel correct or incorrect words; whereas Repeat measures a model's tendency to repeat itself, which is a frequent issue with encoder-decoder models \cite{suzuki2017cutting}. 

On title generation, \textit{rwmd-rank} achieved the best performance in terms of selecting a sentence as the title. In overall, the abstractive systems significantly outperformed the extractive systems, as well as the extractive oracle.  \textit{c2c} and \textit{fconv} performed much better than \textit{lstm}, with a very high rate of overlap. The ROUGE performance of \textit{c2c} and \textit{fconv} is similar, despite the difference of a few R-2 points in favour of \textit{fconv} (that model is evaluated on a subword-level ground truth file, where we observe a slight increase of 1-2 ROUGE points on average due to the conversion).

On abstract generation, the \textit{lead-10} baseline remained tough to beat in terms of ROUGE, and only the extractive systems managed to surpass it by a small margin. All extractive systems achieved similar results, with \textit{rwmd-rank} having a minor edge, while the abstractive \textit{fconv} performed poorly, even though it performed best in terms of METEOR. We observed a much higher repeat rate in the output summaries than the observed $44\%$ average in the original abstracts (Table \ref{table:datasets}). As revealed by the large Repeat standard deviation for \textit{fconv}, some examples are affected by very frequent repetitions. 

\subsection{Qualitative Evaluation}

\begin{figure*}
\centering
\begin{subfigure}[b]{.39\linewidth}
\includegraphics[width=\linewidth]{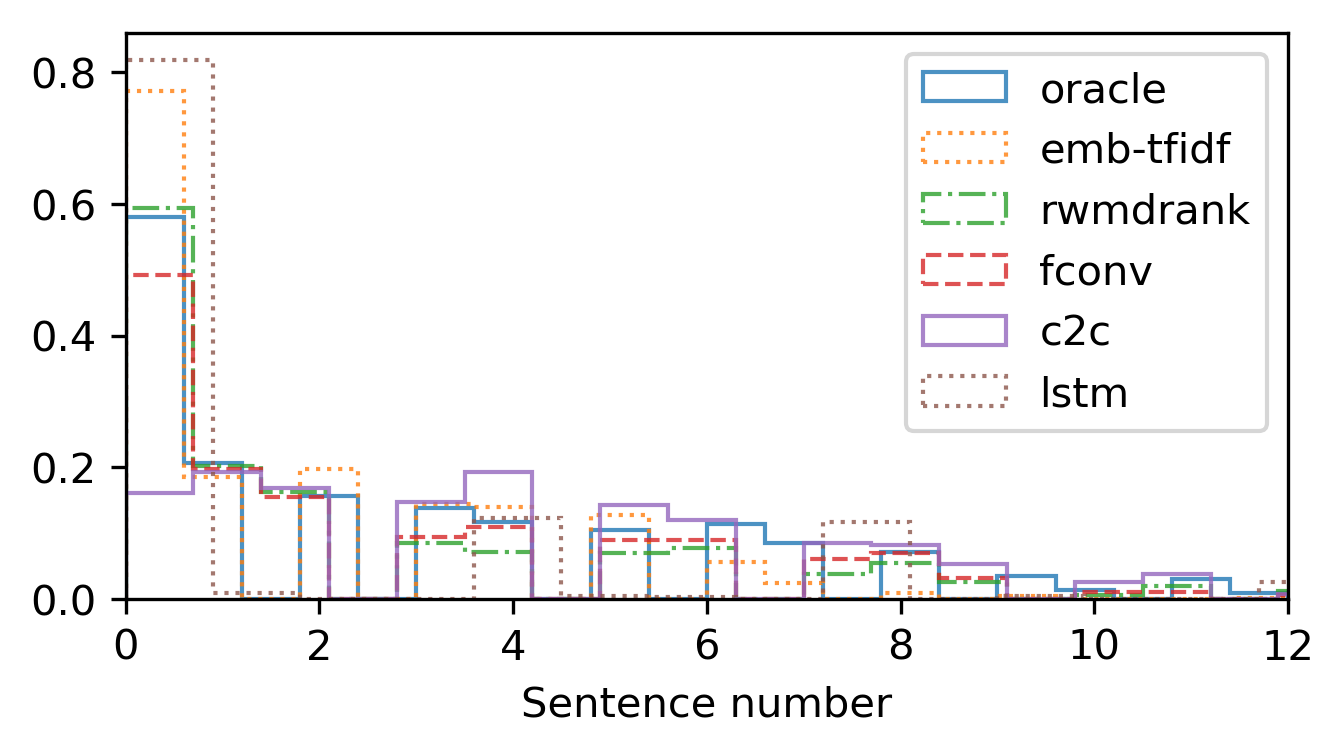}
\caption{}\label{fig:title-hist}
\end{subfigure}
\begin{subfigure}[b]{.41\linewidth}
\includegraphics[width=\linewidth]{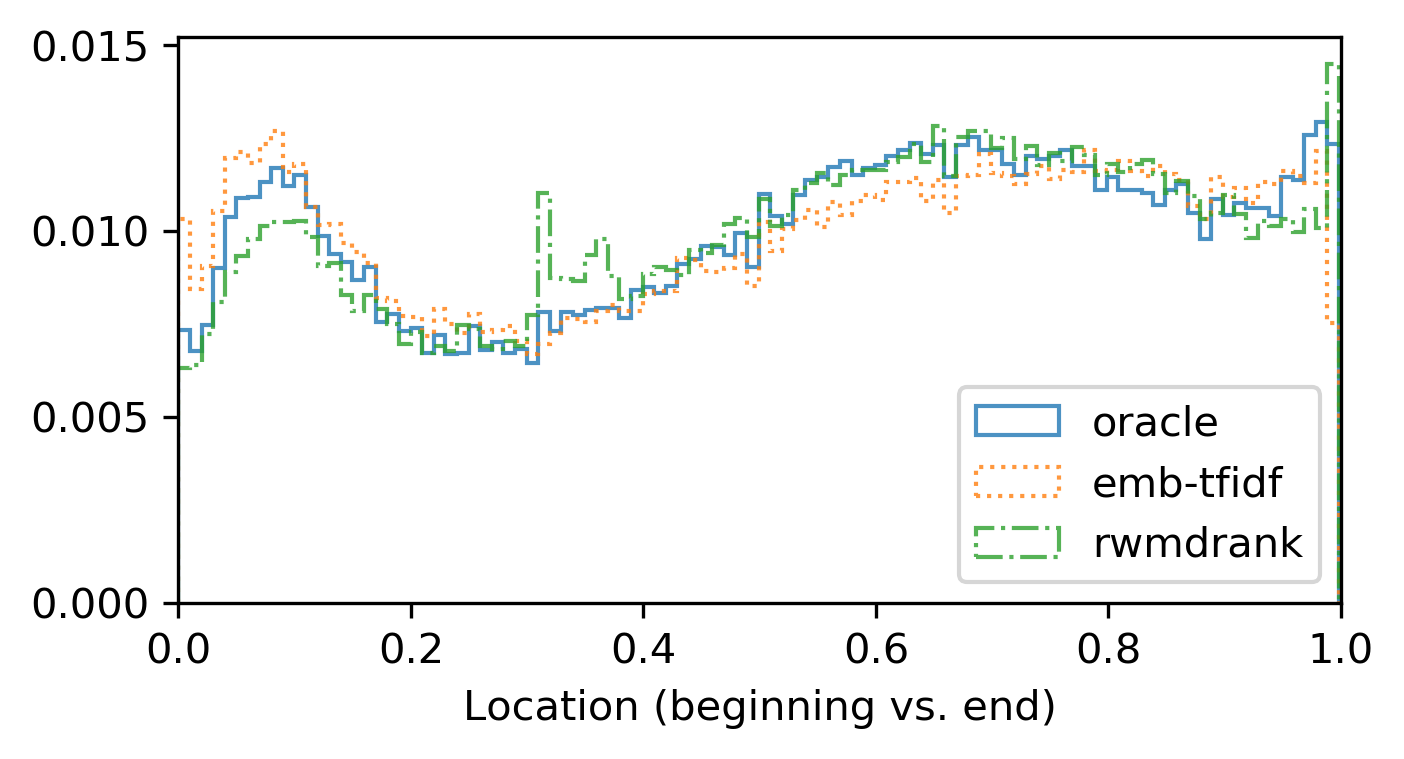}
\caption{}\label{fig:abstract-hist}
\end{subfigure}
\begin{subfigure}[b]{.17\linewidth}
\includegraphics[width=\linewidth]{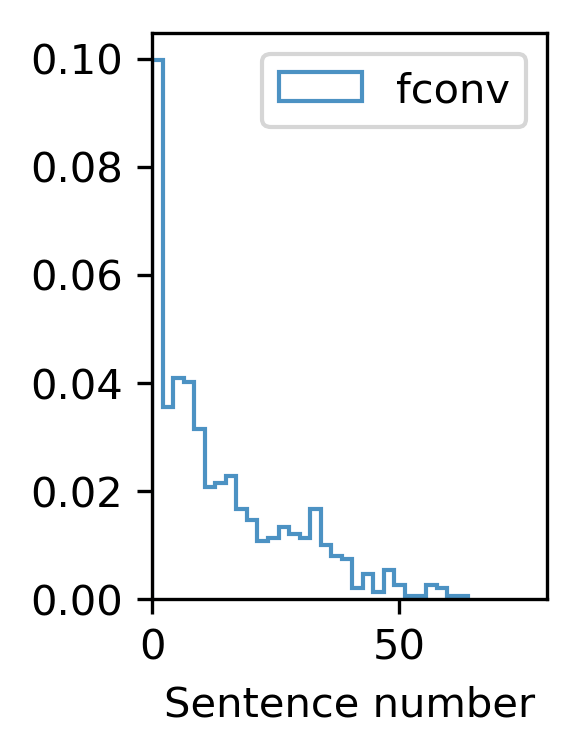}
\caption{}\label{fig:fconv-hist}
\end{subfigure}
\caption{Sentence selection (normalized) histograms computed on the test set, showing the input locations that were most preferred on average by the systems on \textit{title-gen} (a) and \textit{abstract-gen} (b), (c). For (b), we normalize the sentence locations by the length of each paper, to get a better uniform view (there is a large variation in the length of a paper, as shown in Table \ref{table:datasets}). For the abstractive systems, we search for the closest sentences in the input using relaxed word mover's distance (see Section \ref{extractive}).}
\label{fig:histogram}
\end{figure*}

In Tables \ref{table:ex-title-gen} and \ref{table:ex-abstract-gen}, we present two shortened inputs from our \textit{title-gen} and \textit{abstract-gen} test sets, along with original and system-generated summaries. In Figure \ref{fig:histogram}, we show a histogram of the locations of input sentences, that estimates which locations were most preferred on average when producing a summary. 

We observe a large variation in the sentence locations selected by the extractive systems on \textit{title-gen} (Figure \ref{fig:title-hist}), with the first sentence having high importance. Based on our inspection, it is rare that a sentence from the abstract will match the title exactly - the title is also typically shorter than an average sentence from the abstract (Table \ref{table:datasets}). A good title seems to require the selection, combination and paraphrasing of suitable parts from multiple sentences, as also shown by the original titles in our examples. Many of the titles generated by the abstractive systems sound faithful, and at first glance can pass for a title of a scientific paper. The abstractive models are good at discerning important from unimportant content in the abstract, at extracting long phrases, or sometimes whole sentences, and at abstractively combining the information to generate a title. \textit{lstm} is more prone to generate novel words, whereas \textit{c2c} and \textit{fconv} mostly rely on direct copying of content from the abstract, as also indicated by their overlap scores. 

Closer inspection of the titles reveals occasional subtle mistakes: for example, in the first example in Table \ref{table:ex-title-gen}, the \textit{fconv} model incorrectly selected "scopolamine- and cisplatin- induced" which was investigated in the previous work of the authors and is not the main focus of the article. The model also copied the incorrect genus, "mouse" instead of "rat". Sometimes the generated titles sound too general, and fail to communicate the specifics of the paper: in the second example, all models produced "a model of basal ganglia", missing to include the keyword "reinforcement learning": "a model of reinforcement learning in the basal ganglia". These mistakes highlight the complexity of the task, and show that there is still much room for further improvement.

As shown in Figure \ref{fig:abstract-hist}, the introductory and concluding sections are often highly relevant for abstract generation, however relevant content is spread across the entire paper. Interestingly, in the example in Table \ref{table:ex-abstract-gen}, there is a wide range of content that was selected by the extractive systems, with little overlap across systems. For instance, \textit{rwmd-rank} overlaps with \textit{oracle} by 3 sentences, and only by 1 sentence with \textit{emb-tfidf}. 
The outputs of the abstractive \textit{fconv} system on abstract generation are poor in quality, and many of the generated abstracts lack coherent structure and content flow. There is also frequent repetition of entire sentences, as shown by the last sentences produced by \textit{fconv} in Table \ref{table:ex-abstract-gen}. \textit{fconv} also appears to only use the first 60 sentences of the paper to construct the abstract (Figure \ref{fig:fconv-hist}). 

\section{Conclusion}

We evaluated a range of extractive and abstractive neural network-based summarization approaches on two novel datasets constructed from scientific journal articles. While the results for title generation are promising, the models struggled with generating the abstract. This difficulty highlights the necessity for developing novel models capable of efficiently dealing with long input and output sequences, while at the same time preserving the quality of generated sentences. We hope that our datasets will promote more work in this area. A direction to explore in future work is hybrid extractive-abstractive end-to-end approaches that jointly select content and then paraphrase it to produce a summary.

{\fontsize{6}{6}\selectfont
\begin{table*}
\centering
\caption{\small{Examples from the test set of \textit{title-gen}. The outputs of the extractive systems are highlighted as: \hl{oracle}, \hlgreen{tfidf-emb}, \hlcyan{rwmd-rank}. For the abstractive systems, we manually highlighted the text of the concepts that are relevant for the task (errors are highlighted in red).}}
\label{table:ex-title-gen}
\begin{tabularx}{\textwidth}{|X|}
\hline
\small{ 
\textit{Example 1 \cite{giridharan2015schisandrin} Abstract:} 
\hlcyan{Amyloid β (Aβ)-induced neurotoxicity is a major pathological mechanism of Alzheimer’s disease (AD).} \hl{Our previous studies have demonstrated that schisandrin B (Sch B), an antioxidant lignan from Schisandra chinensis, could protect mouse brain against scopolamine- and cisplatin-induced neuronal dysfunction.} In the present study, we examined the {\color{NavyBlue}protective effect of Sch B} against {\color{OliveGreen}intracerebroventricular (ICV)-infused Aβ-induced neuronal dysfunction} in rat cortex and explored the potential mechanism of its action. \hlgreen{Our results showed that 26 days co-administration of Sch B significantly improved the behavioral performance of Aβ (1–40)-infused rats in step-through test.} At the same time, Sch B attenuated Aβ-induced increases in oxidative and nitrosative stresses \textit{(...)}
The aforementioned effects of {\color{NavyBlue}Sch B}  suggest its {\color{NavyBlue}protective role} against {\color{OliveGreen}Aβ-induced neurotoxicity} through intervention in the negative cycle of RAGE-mediated Aβ accumulation during AD patho-physiology.
}
\\ \hline
\small{ \textit{Original title:}
{\color{NavyBlue} schisandrin b ameliorates} {\color{OliveGreen} icv-infused amyloid β induced oxidative stress and neuronal dysfunction} through inhibiting rage / nf-κb / mapk and up-regulating hsp / beclin expression
}
 \\ \hline 
\small{
\textit{lstm:} {\color{NavyBlue}schisandrin b} , an antioxidant lignan from schisandra chinensis , {\color{NavyBlue}protects} against {\color{OliveGreen}amyloid β-induced neurotoxicity}
} 
\\ \hline
\small{
\textit{c2c:} 
{\color{NavyBlue} schisandra chinensis b protects} against {\color{OliveGreen}intracerebroventricular-infused amyloid β induced neuronal dysfunction} in rat cortex 
}
\\ \hline
\small{
\textit{fconf:} 
{\color{NavyBlue} schisandrin b protects} {\color{Red}mouse} brain against {\color{Red}scopolamine- and cisplatin- induced neurotoxicity} in rats 
}
\\ \hline 
\small{\textit{Example 2 \cite{fee2012oculomotor} Abstract:} 
\hlcyan{In its simplest formulation, reinforcement learning is based on the idea that if an action taken in a particular context is followed by a favorable outcome, then, in the same context, the tendency to produce that action should be strengthened, or reinforced.} \textit{(...)}
Recent experiments in the {\color{Purple}songbird} suggest that vocal-related BG circuitry receives two functionally distinct excitatory inputs. \textit{(...)}
\hl{The other is an efference copy of motor commands from a separate cortical brain region that generates vocal variability during learning.} Based on these findings, I propose here a {\color{NavyBlue}general model} of {\color{OliveGreen}vertebrate BG function} that combines context information with a distinct motor efference copy signal. \textit{(...)}
\hlgreen{The model makes testable predictions about the anatomical and functional properties of hypothesized context and efference copy inputs to the striatum from both thalamic and cortical sources.}
}
\\ \hline
\small{ \textit{Original title:}
oculomotor learning revisited : a {\color{NavyBlue}model of reinforcement learning} {\color{OliveGreen} in the basal ganglia} incorporating an efference copy of motor actions . 
}
 \\ \hline 
\small{
\textit{lstm:} a {\color{NavyBlue} model} of basal ganglia function .
} 
\\ \hline
\small{
\textit{c2c:} 
a {\color{NavyBlue}general model} of {\color{Purple}vertebrate} {\color{OliveGreen}basal ganglia function} .
}
\\ \hline
\small{
\textit{fconf:} 
a {\color{NavyBlue}model} of {\color{OliveGreen}basal ganglia function} {\color{Purple}in the songbird} .
}
\\ \hline 
\end{tabularx}
\end{table*}
}

{\fontsize{6}{6}\selectfont
\begin{table*}
\centering
\caption{\small{Two examples from the test set of \textit{abstract-gen}. The outputs of the extractive systems are highlighted as: \hlgreen{tfidf-emb} and \hlcyan{rwmd-rank}, whereas \hlgray{gray} denotes overlap between the two. In \textbf{bold} we mark the content that was selected by the \textit{fconv} system (next page in full), and in \underline{underline} we mark the selection of the \textit{oracle}.}}
\label{table:ex-abstract-gen}
\begin{tabularx}{\textwidth}{|X|}
\multicolumn{1}{l}{} \\
\hline
\small{
\textit{Example 1 \cite{pyysalo2011analysis} Body:} 
\hlgreen{\textbf{In recent years, there has been a significant shift in focus in biomedical information extraction from simple pairwise relations representing associations such as protein-protein interactions (PPI) toward representations that capture typed, structured associations of arbitrary numbers of entities in specific roles, frequently termed event extraction} [1].} Much of this work draws on the GENIA Event corpus \textit{(...)} \ul{This resource served also as the source for the annotations in the first collaborative evaluation of biomedical event extraction methods, the 2009 BioNLP shared task on event extraction (BioNLP ST) [6] as well as for the GENIA subtask of the second task in the series [7, 8].}  \ul{Another recent trend in the domain is a move toward the application of extraction methods to the full scale of the existing literature, with results for various targets covering the entire PubMed literature database of nearly 20 million citations being made available [9, 10, 11, 12].} \hlgreen{\textbf{As event extraction methods} initially developed to target the set of events defined in the GENIA / BioNLP ST corpora are now being applied at PubMed scale, it makes sense to ask how much of the full spectrum of gene/protein associations found there they can maximally cover.} \textit{(...)} By contrast, we will assume that associations not appearing in this data cannot be extracted: as the overwhelming majority of current event extraction methods \textbf{are based on supervised machine learning or hand-crafted rules written with reference to the annotated data}, it reasonable to assume as a first approximation that their coverage of associations not appearing in that data is zero. \underline{\hlcyan{In this study, we seek to characterize the full range of associations of specific genes/proteins described in the literature and}} \underline{\hlcyan{estimate what coverage of these associations event extraction systems relying on currently available resources can}} \underline{\hlcyan{maximally achieve.}} To address these questions, \textbf{it is necessary not only to have an inventory of concepts that (largely) covers the ways in which genes/proteins can be associated, but also to be able to estimate the relative frequency with which these concepts are used to express gene/protein associations} in the literature. \textit{(...)} Here, as we are interested in particular in texts describing associations between two or more gene/protein related entities, \textbf{we apply a focused selection, picking only those individual sentences in which} two or more mentions co-occur. \hlgreen{While this excludes associations in which the entities occur in different sentences, their relative frequency is expected to be low: for example, in the BioNLP ST data, all event participants occurred within a single sentence in 95\% of the targeted biomolecular event statements.} \textit{(...)} \hlcyan{Here, we follow the assumption that when two entities are stated to be associated in some way, the most important words expressing their association will typically be found on the shortest dependency path connecting the two entities (cf.} the shortest path hypothesis of Bunescu and Mooney [30]). The specific dependency representation
\textit{(...)} Table 3 shows the words most frequently occurring on these paths. \hlcyan{This list again suggests an increased focus on words relating to gene/protein associations: expression is the most frequent word on the paths, and binding appears in the top-ranked words.} \textit{(...)}
Finally, to make this pair data consistent with the TPS event spans, tokenization and other features, we aligned the entity annotations of the two corpora. \textit{(...)} \hlgreen{This processing was applied to the BioNLP ST training set, creating a corpus of 6889 entity pairs of which 1119 (16\%) were marked as expressing an association (positive).} \textit{(...)} Evaluation.
\hlgreen{We first evaluated each of the word rankings discussed in the section on Identification of Gene/Protein Associations by comparing the ranked lists of words against the set of single words marked as trigger expressions in the BioNLP ST development data.} \textit{(...)} \hlgreen{To evaluate the capability of the presented approach to identify new expressions of gene/protein associations, we next performed a manual study of candidate words for stating gene/protein associations using the E w ranking.} \textit{(...)} \hlcyan{We then selected the words ranked highest by E w that were not known, grouped by normalized and lemmatized form, and added for reference examples of frequent shortest dependency paths on which any of these words appear (see example in Table 5).} 
\textit{(...)} \ul{If static relations and experimental observations and manipulations are excluded as (arguably) not in scope for event extraction, this estimate suggests that currently available resources for event extraction cover over 90\% of all events involving gene/protein entities in PubMed.} Discussion.
\underline{\hlcyan{We found that out of all gene/protein associations in PubMed, currently existing resources for event extraction are lacking}} \underline{\hlcyan{in coverage of a number of event types such as dissociation, many relatively rare (though biologically important) protein}} \underline{\hlcyan{post-translational modifications, as well as some high-level process types involving genes/proteins such as apoptosis.}} \textit{(...)} \ul{This suggests that for practical applications it may be important to consider also this class of associations.}  \textit{(...)} \hlgreen{While these results are highly encouraging, it must be noted that the approach to identifying gene/protein associations considered here is limited in a number of ways: it excludes associations stated across sentence boundaries and ones for which the shortest path hypothesis does not hold, does not treat multi-word expressions as wholes, ignores ambiguity in implicitly assuming a single sense for each word, and only directly includes associations stated between exactly two entities.} \hlcyan{The approach is also fundamentally limited to associations expressed through specific words and thus blind to e.g. part-of relations implied by statements such as CD14 Sp1-binding site.} \textit{(...)} Conclusions. \underline{\hlgray{We have presented an approach to discovering expressions of gene/protein associations from PubMed based on named}} \underline{\hlgray{entity co-occurrences, shortest dependency paths and an unlexicalized classifier to identify likely statements of gene/protein}} \underline{\hlgray{associations.}} \hlgreen{Drawing on the automatically created full-PubMed annotations of the Turku PubMed-Scale (TPS) corpus and using the BioNLP’09 shared task data to define positive and negative examples of association statements, we distilled an initial set of over 30 million protein mentions into a set of 46,000 unique unlexicalized paths estimated likely to express gene/protein associations. These paths were then used to rank all words in PubMed by the expected number of times they are predicted to express such associations, and 1200 candidate association-expressing words not appearing in the BioNLP’09 shared task data evaluated manually.} \hlcyan{Study of these candidates suggested 18 new event classes for the GENIA ontology and indicated that the majority of statements of gene/protein associations not covered by currently available resources are not statements of biomolecular events but rather statements of static relations or experimental manipulation.} \textit{(...)} \hlcyan{It could thus be assumed that the event types and the specific statements annotated in GENIA would have only modest coverage of all gene/protein association types and statements in PubMed.} \textit{(...)}
}
  \\ \hline
\end{tabularx}
\end{table*}
}

{\fontsize{6}{6}\selectfont
\begin{table*}
\begin{tabularx}{\textwidth}{|X|}
\multicolumn{1}{l}{} \\
\hline
\small{\textit{Example 1 Original abstract:} 
Background: Event extraction following the GENIA Event corpus and BioNLP shared task models has been a considerable focus of recent work in biomedical information extraction. This work includes efforts applying event extraction methods to the entire PubMed literature database, far beyond the narrow subdomains of biomedicine for which annotated resources for extraction method development are available. Results: In the present study, our aim is to estimate the coverage of all statements of gene/protein associations in PubMed that existing resources for event extraction can provide. We base our analysis on a recently released corpus automatically annotated for gene/protein entities and syntactic analyses covering the entire PubMed, and use named entity co-occurrence, shortest dependency paths and an unlexicalized classifier to identify likely statements of gene/protein associations. A set of high-frequency/high-likelihood association statements are then manually analyzed with reference to the GENIA ontology. Conclusions: We present a first estimate of the overall coverage of gene/protein associations provided by existing resources for event extraction. Our results suggest that for event-type associations this coverage may be over 90\%. We also identify several biologically significant associations of genes and proteins that are not addressed by these resources, suggesting directions for further extension of extraction coverage.
}
 \\ \hline
 \small{\textit{Example 1 fconv:} 
background in recent years , there has been a significant shift in focus in biomedical information extraction from simple pairwise relations representing associations such as protein-protein interactions ( ppi ) toward representations that capture typed , structured associations of arbitrary numbers of entities in specific roles , frequently termed event extraction . as event extraction methods are based on supervised machine learning or hand-crafted rules written with reference to the annotated data , it is necessary not only to have an inventory of concepts that ( largely ) covers the ways in which genes / proteins can be associated , but also to be able to estimate the relative frequency with which these concepts are used to express gene / protein associations . results we apply a focused selection , picking king only those individual sentences in which at least one named entity has been tagged . we apply a focused selection , picking only those individual sentences in which at least one named entity has been tagged . we apply a focused selection , picking only those individual sentences in which at least one named entity has been tagged . we apply a probabilistic approach
 } 
  \\ \hline
\small{\textit{Example 2 \cite{kumar2013colocalization} Body:} \underline{\textbf{\hlgray{Both alopecia areata (AA) and vitiligo are autoimmune diseases, and their}}} \underline{\textbf{\hlgray{coexistence in the same patient is not uncommon, as vitiligo has been reported to occur in 4.1\% of patients of AA }}} \underline{\textbf{\hlgray{and is about 4 times more common in patients with AA than in the general population.}}} [1] However, their colocalization over the same site is exceedingly rare, with less than five cases being reported in the literature.[2,3,4] \underline{\hlgray{\textbf{We present a case of a 15-year-old male child who had vitiligo and later developed AA over the} existing lesions of}} \underline{\hlgray{vitiligo over \textbf{face and scalp} and have attempted to elucidate the current understanding of mechanisms of coexistence of}} \underline{\hlgray{these two diseases. }} \hlcyan{A 12-year-old boy presented to the skin outpatient department with history of depigmented areas on the scalp, face, neck, arms and legs for 5 years. He also gave a history of development of patchy loss of hair over some of these lesions for 3 years.} There was no previous history of any trauma or medications. Family history was not relevant. \hlcyan{On examination, there were \textbf{depigmented macules } over the \textbf{scalp}, \textbf{forehead}, \textbf{eyebrows}, \textbf{eyebrows, perioral, preauricular regions}, neck, elbows, hands, feet, shins, \textbf{nose, chin}, hands, knees and feet. Patches of hair loss were seen, limited to some of these depigmented areas over the vertex and occipital region of the scalp and eyebrows [Figure 3]. Other body areas were not affected by patchy hair loss.} \hlgreen{Clinically, the diagnosis of vitiligo with AA was made.} \textit{(..)}
\hlgreen{Additionally, the basal layer of the epidermis was almost devoid of pigment, [Figure 5] confirming the diagnosis of vitiligo over the same site.} \textit{(..)} 
\hlgreen{Both AA and vitiligo are clubbed under the spectrum of autoimmune disorders.} \textit{(..)}
\underline{\hlgreen{Our case lends support to the hypothesis that AA and vitiligo share a common pathogenic pathway including autoimmune}} \underline{\hlgreen{response against some common antigens like those derived from the bulb melanocytes.}}\hlgreen{ Melanocytes-derived peptide antigens released during vitiligo pathogenesis could act as auto-antigens not only for vitiligo, but also for AA, and autoimmune Th-cells against them could also trigger a response against the hair follicle melanocytes, thus, pre-disposing to AA.} \textit{(..text omitted..)} 
}
\\ \hline 
\small{
\textit{Example 2 Original abstract:}
Both alopecia areata (AA) and vitiligo are autoimmune diseases, and their coexistence in the same patient is not uncommon, as vitiligo has been reported to occur in 4.1\% of patients of AA. We present a case of a 15-year-old male child who had vitiligo and later developed AA over the existing lesions of vitiligo over face and scalp and have attempted to elucidate the current understanding of mechanisms of coexistence of these two diseases. Our case lends support to the hypothesis that AA and vitiligo share a common pathogenic pathway including autoimmune response against some common antigens like those derived from the bulb melanocytes. Stimulation of proinflammatory T-helper(Th)-1 cell mediated immunological response or inactivation of a suppressor T-cell mediated response could be the common underlying phenomenon. However, the striking rarity of colocalization of these two diseases has led to the recent debate over current understanding of their pathogenesis, and whether this association is merely a coincidence. As both AA and vitiligo are frequent and chronic dermatological disorders, it is of utmost importance to gain more understanding into their pathogenic mechanisms so that more definitive treatment modalities may be devised and the quality of life of these patients can be improved.}
\\ \hline
\small{\textit{Example 2 fconv:} alopecia areata ( aa ) and vitiligo are autoimmune diseases , and their coexistence in the same patient is not uncommon , as vitiligo has been reported to occur in 4.1 \% of patients of aa and is about 4 times more common in patients with aa than in the general population . we present a case of a 15- year-old male child who had vitiligo and later developed aa over the scalp , forehead , eyebrows , eyebrows , perioral , preauricular regions , nose and chin depigmented macules over the scalp , forehead , eyebrows , eyebrows , perioral , preauricular regions , nose and chin depigmented macules over the scalp , forehead , eyebrows , periorbital , perioral , preauricular regions , nose and chin depigmented macules over the scalp , forehead , eyebrows , periorbital , perioral , preauricular regions , nose and chin depigmented macules over the scalp , forehead , eyebrows , periorbital}
\\ \hline
\end{tabularx}
\end{table*}
}

\section{Acknowledgments}

We thank the reviewers for their useful comments, and NVIDIA for the donation of a TITAN X graphics card. 

\section{References}


\begin{thebibliography}{}

\bibitem[\protect\citename{Bahdanau \bgroup et al.\egroup
  }2014]{bahdanau2014neural}
Bahdanau, D., Cho, K., and Bengio, Y.
\newblock (2014).
\newblock Neural machine translation by jointly learning to align and
  translate.
\newblock {\em arXiv preprint arXiv:1409.0473}.

\bibitem[\protect\citename{Biber and Gray}2010]{biber2010challenging}
Biber, D. and Gray, B.
\newblock (2010).
\newblock Challenging stereotypes about academic writing: Complexity,
  elaboration, explicitness.
\newblock {\em Journal of English for Academic Purposes}, 9(1):2--20.

\bibitem[\protect\citename{Brokos \bgroup et al.\egroup }2016]{brokos2016using}
Brokos, G.-I., Malakasiotis, P., and Androutsopoulos, I.
\newblock (2016).
\newblock Using centroids of word embeddings and word mover's distance for
  biomedical document retrieval in question answering.
\newblock {\em arXiv preprint arXiv:1608.03905}.

\bibitem[\protect\citename{Chiu \bgroup et al.\egroup }2016]{chiu2016train}
Chiu, B., Crichton, G., Korhonen, A., and Pyysalo, S.
\newblock (2016).
\newblock How to train good word embeddings for biomedical nlp.
\newblock {\em Proceedings of BioNLP16}, page 166.

\bibitem[\protect\citename{Cho \bgroup et al.\egroup }2014]{cho2014learning}
Cho, K., Van~Merri{\"e}nboer, B., Gulcehre, C., Bahdanau, D., Bougares, F.,
  Schwenk, H., and Bengio, Y.
\newblock (2014).
\newblock Learning phrase representations using rnn encoder-decoder for
  statistical machine translation.
\newblock {\em arXiv preprint arXiv:1406.1078}.

\bibitem[\protect\citename{Cho \bgroup et al.\egroup }2015]{cho2015describing}
Cho, K., Courville, A., and Bengio, Y.
\newblock (2015).
\newblock Describing multimedia content using attention-based encoder--decoder
  networks.
\newblock {\em arXiv preprint arXiv:1507.01053}.

\bibitem[\protect\citename{Collins \bgroup et al.\egroup
  }2017]{collins2017supervised}
Collins, E., Augenstein, I., and Riedel, S.
\newblock (2017).
\newblock A supervised approach to extractive summarisation of scientific
  papers.
\newblock {\em arXiv preprint arXiv:1706.03946}.

\bibitem[\protect\citename{Denkowski and
  Lavie}2014]{denkowski:lavie:meteor-wmt:2014}
Denkowski, M. and Lavie, A.
\newblock (2014).
\newblock Meteor universal: Language specific translation evaluation for any
  target language.
\newblock In {\em Proceedings of the EACL 2014 Workshop on Statistical Machine
  Translation}.

\bibitem[\protect\citename{Dietterich \bgroup et al.\egroup
  }2008]{dietterich2008structured}
Dietterich, T.~G., Domingos, P., Getoor, L., Muggleton, S., and Tadepalli, P.
\newblock (2008).
\newblock Structured machine learning: the next ten years.
\newblock {\em Machine Learning}, 73(1):3--23.

\bibitem[\protect\citename{Erkan and Radev}2004]{erkan2004lexrank}
Erkan, G. and Radev, D.~R.
\newblock (2004).
\newblock Lexrank: Graph-based lexical centrality as salience in text
  summarization.
\newblock {\em Journal of Artificial Intelligence Research}, 22:457--479.

\bibitem[\protect\citename{Fee}2012]{fee2012oculomotor}
Fee, M.~S.
\newblock (2012).
\newblock Oculomotor learning revisited: a model of reinforcement learning in
  the basal ganglia incorporating an efference copy of motor actions.
\newblock {\em Frontiers in neural circuits}, 6:38.

\bibitem[\protect\citename{Gehring \bgroup et al.\egroup
  }2017]{gehring2017convolutional}
Gehring, J., Auli, M., Grangier, D., Yarats, D., and Dauphin, Y.~N.
\newblock (2017).
\newblock Convolutional sequence to sequence learning.
\newblock {\em arXiv preprint arXiv:1705.03122}.

\bibitem[\protect\citename{Giridharan \bgroup et al.\egroup
  }2015]{giridharan2015schisandrin}
Giridharan, V.~V., Thandavarayan, R.~A., Arumugam, S., Mizuno, M., Nawa, H.,
  Suzuki, K., Ko, K.~M., Krishnamurthy, P., Watanabe, K., and Konishi, T.
\newblock (2015).
\newblock Schisandrin b ameliorates icv-infused amyloid $\beta$ induced
  oxidative stress and neuronal dysfunction through inhibiting
  rage/nf-$\kappa$b/mapk and up-regulating hsp/beclin expression.
\newblock {\em PLoS One}, 10(11):e0142483.

\bibitem[\protect\citename{Hunter and Cohen}2006]{hunter2006biomedical}
Hunter, L. and Cohen, K.~B.
\newblock (2006).
\newblock Biomedical language processing: what's beyond pubmed?
\newblock {\em Molecular cell}, 21(5):589--594.

\bibitem[\protect\citename{Kalchbrenner \bgroup et al.\egroup
  }2016]{Kalchbrenner2016Bytenet}
Kalchbrenner, N., Espeholt, L., Simonyan, K., van~den Oord, A., Graves, A., and
  Kavukcuoglu, K.
\newblock (2016).
\newblock Neural machine translation in linear time.
\newblock {\em CoRR}, abs/1610.10099.

\bibitem[\protect\citename{Kim \bgroup et al.\egroup }2016]{kim2016towards}
Kim, M., Singh, M.~D., and Lee, M.
\newblock (2016).
\newblock Towards abstraction from extraction: Multiple timescale gated
  recurrent unit for summarization.
\newblock {\em arXiv preprint arXiv:1607.00718}.

\bibitem[\protect\citename{Kumar \bgroup et al.\egroup
  }2013]{kumar2013colocalization}
Kumar, S., Mittal, J., and Mahajan, B.
\newblock (2013).
\newblock Colocalization of vitiligo and alopecia areata: coincidence or
  consequence?
\newblock {\em International journal of trichology}, 5(1):50.

\bibitem[\protect\citename{Kusner \bgroup et al.\egroup }2015]{kusner2015word}
Kusner, M.~J., Sun, Y., Kolkin, N.~I., Weinberger, K.~Q., et~al.
\newblock (2015).
\newblock From word embeddings to document distances.
\newblock In {\em ICML}, volume~15, pages 957--966.

\bibitem[\protect\citename{LeCun \bgroup et al.\egroup }1998]{Lecun1998CNN}
LeCun, Y., Bottou, L., Bengio, Y., and Haffner, P.
\newblock (1998).
\newblock Gradient-based learning applied to document recognition.
\newblock {\em Proceedings of the IEEE}, 86(11):2278--2324, Nov.

\bibitem[\protect\citename{Lee \bgroup et al.\egroup }2016]{lee2016fully}
Lee, J., Cho, K., and Hofmann, T.
\newblock (2016).
\newblock Fully character-level neural machine translation without explicit
  segmentation.
\newblock {\em arXiv preprint arXiv:1610.03017}.

\bibitem[\protect\citename{Lin}2004]{lin2004rouge}
Lin, C.-Y.
\newblock (2004).
\newblock Rouge: A package for automatic evaluation of summaries.
\newblock In {\em Text summarization branches out: Proceedings of the ACL-04
  workshop}, volume~8.

\bibitem[\protect\citename{Lloret \bgroup et al.\egroup
  }2013]{lloret2013compendium}
Lloret, E., Rom{\'a}-Ferri, M.~T., and Palomar, M.
\newblock (2013).
\newblock Compendium: A text summarization system for generating abstracts of
  research papers.
\newblock {\em Data \& Knowledge Engineering}, 88:164--175.

\bibitem[\protect\citename{Mikolov \bgroup et al.\egroup
  }2013]{mikolov2013efficient}
Mikolov, T., Chen, K., Corrado, G., and Dean, J.
\newblock (2013).
\newblock Efficient estimation of word representations in vector space.
\newblock {\em arXiv preprint arXiv:1301.3781}.

\bibitem[\protect\citename{Nallapati \bgroup et al.\egroup
  }2016]{nallapati2016sequence}
Nallapati, R., Zhou, B., Gul{\c{c}}ehre, {\c{C}}., and Xiang, B.
\newblock (2016).
\newblock Abstractive text summarization using sequence-to-sequence rnns and
  beyond.
\newblock {\em arXiv preprint arXiv:1602.06023}.

\bibitem[\protect\citename{Nenkova \bgroup et al.\egroup
  }2011]{nenkova2011automatic}
Nenkova, A., Maskey, S., and Liu, Y.
\newblock (2011).
\newblock Automatic summarization.
\newblock In {\em Proceedings of the 49th Annual Meeting of the Association for
  Computational Linguistics: Tutorial Abstracts of ACL 2011}, page~3.
  Association for Computational Linguistics.

\bibitem[\protect\citename{Page \bgroup et al.\egroup }1999]{page1999pagerank}
Page, L., Brin, S., Motwani, R., and Winograd, T.
\newblock (1999).
\newblock The pagerank citation ranking: Bringing order to the web.
\newblock Technical report, Stanford InfoLab.

\bibitem[\protect\citename{Pyysalo \bgroup et al.\egroup
  }2011]{pyysalo2011analysis}
Pyysalo, S., Ohta, T., and Tsujii, J.
\newblock (2011).
\newblock An analysis of gene/protein associations at pubmed scale.
\newblock {\em Journal of biomedical semantics}, 2(5):S5.

\bibitem[\protect\citename{Radev \bgroup et al.\egroup
  }2004]{radev2004centroid}
Radev, D.~R., Jing, H., Sty{\'s}, M., and Tam, D.
\newblock (2004).
\newblock Centroid-based summarization of multiple documents.
\newblock {\em Information Processing \& Management}, 40(6):919--938.

\bibitem[\protect\citename{Rossiello \bgroup et al.\egroup
  }2017]{rossiello2017centroid}
Rossiello, G., Basile, P., and Semeraro, G.
\newblock (2017).
\newblock Centroid-based text summarization through compositionality of word
  embeddings.
\newblock {\em MultiLing 2017}, page~12.

\bibitem[\protect\citename{Rush \bgroup et al.\egroup }2015]{rushneural}
Rush, A.~M., Chopra, S., and Weston, J.
\newblock (2015).
\newblock A neural attention model for abstractive sentence summarization.
\newblock {\em arXiv preprint arXiv:1509.00685}.

\bibitem[\protect\citename{Sennrich \bgroup et al.\egroup
  }2015]{sennrich2015neural}
Sennrich, R., Haddow, B., and Birch, A.
\newblock (2015).
\newblock Neural machine translation of rare words with subword units.
\newblock {\em arXiv preprint arXiv:1508.07909}.

\bibitem[\protect\citename{Sutskever \bgroup et al.\egroup
  }2014]{sutskever2014sequence}
Sutskever, I., Vinyals, O., and Le, Q.~V.
\newblock (2014).
\newblock Sequence to sequence learning with neural networks.
\newblock In {\em Advances in neural information processing systems}, pages
  3104--3112.

\bibitem[\protect\citename{Suzuki and Nagata}2017]{suzuki2017cutting}
Suzuki, J. and Nagata, M.
\newblock (2017).
\newblock Cutting-off redundant repeating generations for neural abstractive
  summarization.
\newblock In {\em Proceedings of the 15th Conference of the European Chapter of
  the Association for Computational Linguistics: Volume 2, Short Papers},
  volume~2, pages 291--297.

\bibitem[\protect\citename{Teufel and Moens}2002]{teufel2002summarizing}
Teufel, S. and Moens, M.
\newblock (2002).
\newblock Summarizing scientific articles: experiments with relevance and
  rhetorical status.
\newblock {\em Computational linguistics}, 28(4):409--445.

\end{thebibliography}

\appendix

\end{document}